\PassOptionsToPackage{table,xcdraw}{xcolor}
\documentclass{article}
\usepackage{arxiv}

\usepackage[utf8]{inputenc} 
\usepackage[T1]{fontenc}    
\usepackage{hyperref}       
\usepackage{url}            
\usepackage{booktabs}       
\usepackage{amsfonts}       
\usepackage{nicefrac}       
\usepackage{microtype}      
\usepackage{lipsum}		
\usepackage{graphicx}
\usepackage{natbib}
\usepackage{doi}

\usepackage{amsmath}
\usepackage{xcolor}
\usepackage{subfig}
\usepackage{caption}
\usepackage{multirow}

\title{Can Language Models Capture Graph Semantics? From Graphs to Language Model and Vice-Versa}

\author{{\hspace{1mm}Tarun Garg} \\
	BITS Pilani \\
	\texttt{f20160450h@alumni.bits-pilani.ac.in} \\
	\And
	{\hspace{1mm}Kaushik Roy} \\
	AI Institute, University of South Carolina \\
	\texttt{kaushikr@email.sc.edu} \\
	\And
	{\hspace{1mm}Amit Sheth} \\
	AI Institute, University of South Carolina \\
	\texttt{amit@sc.edu} \\
}

\date{}

\renewcommand{\headeright}{}
\renewcommand{\undertitle}{}
\renewcommand{\shorttitle}{}

\hypersetup{
pdftitle={Can Language Models Capture Graph Semantics? From Graphs to Language Model and Vice-Versa},
pdfsubject={q-bio.NC, q-bio.QM},
pdfauthor={Tarun Garg, Kaushik Roy, Amit Sheth},
pdfkeywords={Knowledge Graph, Graph Neural Networks, Transformers},
}

\begin{document}
\maketitle

\begin{abstract}
   Knowledge Graphs are a great resource to capture semantic knowledge in terms of entities and relationships between the entities. However, current deep learning models takes as input distributed representations or vectors. Thus, the graph is compressed in a vectorized representation. We conduct a study to examine if the deep learning model can compress a graph and then output the same graph with most of the semantics intact. Our experiments show that Transformer models are not able to express the full semantics of the input knowledge graph. We find that this is due to the disparity between the directed, relationship and type based information contained in a Knowledge Graph and the fully connected token-token undirected graphical interpretation of the Transformer Attention matrix. 
\end{abstract}



\maketitle

\section{Introduction}
Natural Language is a widely researched domain focusing on interpreting and processing human language for better human-machine interaction. Taking insights from Natural Language is challenging as most data is unstructured and cannot be fed directly into the machine. Models like BERT \textbf{\cite{devlin2018bert}}, GPT \textbf{\cite{brown2020language}} are increasingly being used to solve major Natural Language problems. Question Answering, Machine translation, and Information retrieval are examples that use these underlying algorithms. In Natural Language research, various forms of embedding like Glove \textbf{\cite{pennington2014glove}}, Word2Vec \textbf{\cite{mikolov2013efficient}}, TF-IDF \textbf{\cite{ramos2003using}} are used to vectorize the textual data. The vector embeddings are then passed over as an input to the Machine learning Models.

Knowledge Graphs are an emerging form of Knowledge capture that represents structured semantic knowledge widely available on the web. Knowledge Graphs is a form of a Graphical Database incorporating domain knowledge resulting in scalable and efficient usability. KGs have been used in the past with Deep Networks to develop algorithms such as Graph Attention Networks  \textbf{\cite{velivckovic2017graph}}, OpenKE \textbf{\cite{han2018openke}}, Graph Convolution Transformer \textbf{\cite{choi2019graph}} which are guided by the structured knowledge to make better predictions.

Use of Knowledge Graphs in language models had been done in the past for various applications like QA-GNN \textbf{\cite{yasunaga2021qa}}, Entity2rec \textbf{\cite{palumbo2017entity2rec}}. They have successfully managed to use KG to improve the performance of Language models. Despite the improved performance, as such methods compress the KG into vectorized representation, these representations haven't yet been demonstrably shown to capture the full semantics information in KG.\textbf{\cite{swamy2021interpreting}}, \textbf{\cite{jain2021embeddings}}.

Considering many applications of Knowledge Graph in Natural Language, we chose to conduct a study to figure out whether Language Models have the semantic capability to store Knowledge Graphs? Here we have performed some experiments to understand it and provided insights and detailed discussion on the possible shortcomings. We have tried to provide adequate pointers to allow further research on this.

\section{Background}

\subsection{Knowledge Graphs}

As mentioned above, Knowledge graphs are Graphical Database being widely used in Natural Language to improve the existing models through techniques such as Knowledge infusion \textbf{\cite{sheth2021knowledge}}, \textbf{\cite{kursuncu2019knowledge}}, \textbf{\cite{su2021cokebert}}.
The use of KG is attractive because of the wide variety of different kinds of knowledge that they capture online. Some of the widely used KGs available on the web are as follows:
\begin{itemize}
\item Wikidata \textbf{\cite{vrandevcic2014wikidata}} is one of the largest, freely available Knowledge Graph, which organizes data in a structured graphical format for better information extraction. It is a multilingual data prepared majorly by Wikipedia and other Wikimedia sister projects.  Wikidata editors maintain the data, and the items are encoded with a unique key starting with Q and followed by a number.
\item ConceptNet is another freely-available semantic network that stores data relating to the words people use and their semantic meaning. The data is prepared from various crowd-sourced projects. This network can be used to help computers understand words from the natural language humans commonly use by creating word embeddings. \textbf{\cite{speer2017conceptnet}}.
\item DBpedia is a crowd-sourced initiative to extract structured content from the information present in various Wikipedia resources. DBpedia is multilingual and represents actual community agreement. It regularly evolves as the content on Wikipedia is updated \textbf{\cite{auer2007dbpedia}}. 
\end{itemize}

\subsection{Language Models}

Language models use deep learning methods to determine the probability of a given sequence of words occurring in a sentence \textbf{\cite{lavrenko2017relevance}}. Language models started with sequence models using RNNs \textbf{\cite{cho2014learning}}, LSTMs \textbf{\cite{hochreiter1997long}}, and GRUs \textbf{\cite{cho2014learning}}. Attention models were discovered that could help the model \textit{focus} on particular parts of the input. Transformer models utilize self attention to learn word and token level correlations while performing predictions \textbf{\cite{vaswani2017attention}}. Recent research has used transformer technology to combine graphs and language models to predict the output.

\subsection{Graph Neural Networks - GNN}

Graph Neural Networks \textbf{\cite{scarselli2008graph}} is a differentiable message passing method that aggregates information from neighbouring nodes into node representations using convolution operations. However, they do not consider relationship information. R-GCN( Relational Graph Convolution Networks) \textbf{\cite{schlichtkrull2018modeling}} attempt to incorporate relationship information on the links enforcing sparsity and regularization constraints while constructing node representations. However, real-world knowledge graphs have few relationships and thus require modulation of a graph node network that does not contain relations with a graph that contains few relations between the same nodes. Thus, we employ the Graph Convolutional Transformer \textbf{\cite{choi2019graph}} to achieve this modulation.

\subsection{Graph Convolution Transformer - GCT}

Graph Convolution Transformer \textbf{\cite{choi2019graph}} uses the prior probabilities calculated from the domain knowledge graph to train the model. The model architecture consists of 3 transformer stacks with 2 feed forward layers. eICU Collaborative Research Database \textbf{\cite{pollard2018eicu}} consisting of anonymized medical details of patients is used to train the model. 

GCT relies on the intuition that the Attention matrix in a Transformer can be thought of as a fully connected weighted graph among tokens. Thus, allowing for integration between the two graphical structures (Knowledge Graph and the Attention matrix graph). However, the nature of the graphs are very different. One is a directed graph with relationships and schema (type) information and the other is a fully connected weighted undirected graph. As we will see in our experiments, this disparity will be an impediment in allowing adequate semantic knowledge capture from the Knowledge Graph by the Transformer architecture.

GCT working relies on 2 major components: i) KL-divergence ii) negative log-likelihood. KL Divergence helps preserve the domain knowledge, and log-likelihood helps find new relations for making predictions. Here we have used GCT and tried some modifications to quantify the domain knowledge captured in the Language Model.

\subsection{Language models are open knowledge graphs}
Previous models have tried to illustrate that language models do capture information present in KG. For example one such methodology to generate Knowledge Graph from Language Models was proposed in 2020  \textbf{\cite{wang2020language}}, which incorporates the concept of Conditional Probability. Here the researchers have proposed that the attention matrix can be used as the conditional probability, which can further be used to construct a knowledge graph. Attention weights denote the correlation of one word with another which can be used to label the corresponding entities and relations, leading to the generation of a new Knowledge Graph.


\begin{figure}
    \centering
    \subfloat[Example path followed to generate triple]{\label{fig:1(a)}\includegraphics{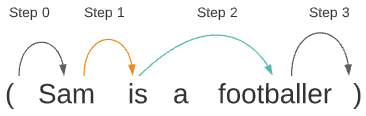}} \\
    \vspace{5mm} 
    \subfloat[Procedure to generate a triple for the given context]{\label{fig:1(b)}
    \begin{tabular}{|c | c | c | c |} 
    \hline
    Step & Action & Possible intermediates & Match\\
    \hline
    0 & START & (Sam, & 0\\
    \hline
    1 & YIELD & (Sam,is, & 0.4\\
    \hline
    2 & YIELD & (Sam,is,footballer & 0.9\\
    \hline
    3 & STOP & (Sam,is,footballer) & 0.9\\
    \hline
    \end{tabular}}\\
    \vspace{5mm}
    \subfloat[Resulting attention matrix from the sample example]{\label{fig:1(c)}
    \begin{tabular}{ c  c | c | c | c | c |} 
    \multicolumn{2}{c} {} & \multicolumn{4}{c} {Query}\\
    \multicolumn{2}{c} {} &  \multicolumn{1}{c}{Sam} & \multicolumn{1}{c}{is} & \multicolumn{1}{c}{a} & \multicolumn{1}{c}{footballer}\\
    \cline{3-6}
    \multirow{4}{1em}{Key} & Sam & x & x & x & x\\\cline{3-6}
    & is & \cellcolor{orange!50} 0.4 & x & x & x\\\cline{3-6}
    & a & 0.1 & 0.2 & x & x\\ \cline{3-6}
    & footballer & 0.2 & \cellcolor{green!50} 0.5 & 0.1 & x\\\cline{3-6}
    \end{tabular}}
    \caption{An illustration showing the procedure to generate the triple }
    \label{fig:triplet_generation}
\end{figure}

In the sample matrix (Figure \ref{fig:triplet_generation}) \footnote{Figures inspired by paper \cite{wang2020language} }, the author has tagged an entity and predicted the next terms of the triple based on conditional probability extracted from the attention matrix. In the example shown, the algorithms start with the word ‘Sam’. Corresponding to ‘Sam’, the word with maximum attention is ‘is’ and similarly ‘footballer’ will be the next word corresponding to ‘is’.

However, there are various restrictions on how this method this method generates the KG. Notably all KG triples are not forward directional and schema level information that is essential to disambiguate entities in different context is not at captured. This schema-level information that contains various contextual relationships represents the heart of the semantics in the KG required for reasoning.

\section{Experiments}
The question remains can we generate the full semantics of a Knowledge Graph from Language Models?
We test the hypothesis that does the Graph Convolution Transformer \textbf{\cite{choi2019graph}} has enough expressivity to encode the causal context in the knowledge graph. If the predicted structure matches the knowledge graph that is encoded in the Graph Convolution Transformer (GCT), the hypothesis is confirmed. If a model such as GCT can capture a specific context such as causal, we may hope that other context representing the full semantics of a KG can be captured in future Language Models.

GCT working involves reading data across various files including diagnosis, treatment details and are grouped based on patient ID. Conditional probabilities are then calculated for Diagnosis-procedure and procedure-diagnosis pairs. The grouped data is then stored as TfRecord, which is then processed by the training module.

Conditional probabilities are generated based on the occurrences of the terms together for a particular patient. These conditional probabilities are marked as prior probabilities used as the non-trainable attention weights of the first transformer stack. Further layers are trained using KL divergence and sigmoid cross-entropy as the loss function to preserve the domain knowledge by penalizing the difference between the attention weights of consecutive layers. We will be using the attention weights generated in the final transformer stack to develop a new graph.

\begin{figure*}
    \centering
    \includegraphics[width=\textwidth]{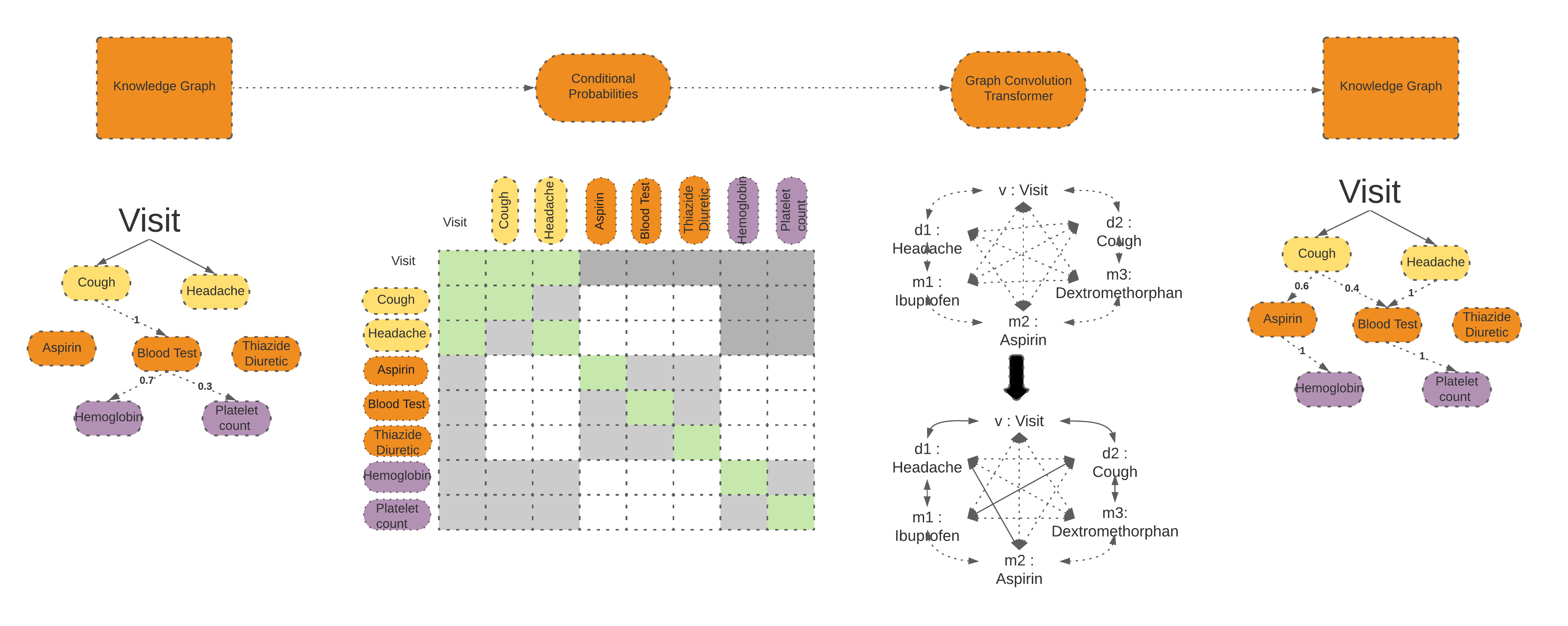}
    \caption{ (a) The first transition transition talks about calculating conditional probabilities based on the occurrence of entity relation pairs. This results in the prior attention matrix.(b) Second transition shows the relational mapping of the context through Graph Convolution Transformer. The darker edged signifies strong relationship between the linked entities. (c) Attention matrix is then used to reproduce the Knowledge Graph based on the probabilities}
    \label{fig:flowchart}
\end{figure*}

We encode the graph structure into the graph neural network through a convolution operation to modulate the graph node representations (GCT). Next, we tried to recover the graph structure by predicting the graph structure from the language model (Language Models are open knowledge graphs) (see \textbf{Figure} \ref{fig:flowchart}).

To understand the importance of GCT architecture, various changes were made in architecture of GCT to measure the effect of each layer. Firstly, condition probabilities of each procedure-disease and disease-procedure relation are calculated and stored separately for each patient. While passing the data to Graph Convolution Transformer. The condition probabilities are pushed as the first attention matrix. KL Divergence is used further along with cross entropy loss to train the model. Note that the key difference from the original GCT paper is that we impose the KL divergence minimization at every transformer block.

KL Divergence Formula
\begin{equation}
\setlength{\jot}{10pt}
    \begin{aligned}
        Define \hat{A}^{(j)} := softmax(\frac{Q^{(j)}K^{(j)T}}{\sqrt{d}} + M) \\
    \end{aligned}
\end{equation}
\begin{equation}
    \setlength{\jot}{10pt}
        \begin{aligned}
        Self-Attention : \\
        C^{(j)} = MLP^{(j)}(PC^{(j-1)}W_v^{(j)}) when j= 1, \\
        C^{(j)} = MLP^{(j)}(\hat{A}^{(j)}C^{(j-1)}W_V^{(j)}) when j> 1 \\
    \end{aligned}
\end{equation}
\begin{equation}
    \setlength{\jot}{10pt}
    \begin{aligned}
        Regularization : \\
        L_{reg}^{(j)} = D_{KL}(P||\hat{A}^{(j)}) when j=1, \\
        L_{reg}^{(j)} = D_{KL}(\hat{A} ^ {(j-1)} || \hat{A} ^ {(j)}) when j>1 \\
    \end{aligned}
\end{equation}
\begin{equation}
    \setlength{\jot}{10pt}
    \begin{aligned}
        Orginal Loss Function : \\
        L = L_{(pred)} + \lambda\sum_{j} L_{reg}^{(j)} \\
    \end{aligned}
\end{equation}
\begin{equation}
    \setlength{\jot}{10pt}
    \begin{aligned}
        Modified Loss Function : \\
        L = \lambda\sum_{j} L_{reg}^{(j)}\\
    \end{aligned}
\end{equation}

To experiment the impact of KL-divergence \textbf{\cite{joyce2011kullback}} on Language Models, we made a change in the loss function of GCT. We removed the cross entropy loss from loss function and allowed the model to train on eICU data \textbf{\cite{pollard2018eicu}} with only KL-divergence. Our hypothesis in this experiment was that with only KL-divergence, the loss between layers should significantly decrease as it will try to preserve maximum possible information. At the same time AUC-PR and AUC-ROC should decrease significantly because it is no more using feedback from the labels. After training with new loss function, we tried reproducing the original knowledge graph.

\section{Results}

\begin{table*}[h!]
\begin{center}
 \begin{tabular}{|c c c c | c c c c|} 
 \hline
 \multicolumn{4} {|c|} {\textbf{Modified Loss function}} & \multicolumn{4} {|c|} {\textbf{Original Loss function}}\\
  \hline
 Steps & AUC-PR & AUC-ROC & loss & Steps & AUC-PR & AUC-ROC & loss\\ [0.5ex] 
 \hline\hline
100 & 0.091 & 0.354 & 0.113 & 100 & 0.136 & 0.569 & 0.817\\ \hline
200 & 0.082 & 0.28 & 0.112 & 200 & 0.122 & 0.517 & 1.09\\ \hline
300 & 0.081 & 0.277 & 0.112 & 300 & 0.148 & 0.599 & 1.013\\ \hline
400 & 0.079 & 0.254 & 0.113 & 400 & 0.127 & 0.519 & 1.129\\ \hline
500 & 0.078 & 0.236 & 0.112 & 500 & 0.118 & 0.497 & 1.297\\ \hline
600 & 0.081 & 0.273 & 0.112 & 600 & 0.122 & 0.51 & 1.29\\ \hline
700 & 0.079 & 0.249 & 0.112 & 700 & 0.139 & 0.576 & 1.345\\ \hline
800 & 0.079 & 0.255 & 0.112 & 800 & 0.109 & 0.464 & 1.128\\ \hline
900 & 0.081 & 0.273 & 0.112 & 900 & 0.097 & 0.408 & 1.302\\ \hline
1000 & 0.08 & 0.264 & 0.112 & 1000 & 0.102 & 0.439 & 1.525\\ \hline
1100 & 0.078 & 0.246 & 0.112 & 1100 & 0.105 & 0.449 & 1.311\\ \hline
1200 & 0.079 & 0.251 & 0.112 & 1200 & 0.109 & 0.466 & 1.314\\ \hline
1300 & 0.077 & 0.239 & 0.112 & 1300 & 0.097 & 0.408 & 1.449\\ \hline
1400 & 0.081 & 0.278 & 0.112 & 1400 & 0.099 & 0.418 & 1.505\\ \hline
1500 & 0.08 & 0.268 & 0.112 & 1500 & 0.105 & 0.449 & 1.557\\ \hline
1600 & 0.083 & 0.299 & 0.112 & 1600 & 0.105 & 0.449 & 1.568\\ \hline
1700 & 0.083 & 0.3 & 0.112 & 1700 & 0.153 & 0.608 & 0.941\\ \hline
1800 & 0.084 & 0.313 & 0.112 & 1800 & 0.125 & 0.507 & 1.14\\ \hline
1900 & 0.086 & 0.322 & 0.112 & 1900 & 0.14 & 0.534 & 1.407\\ \hline
2000 & 0.086 & 0.328 & 0.112 & 2000 & 0.145 & 0.548 & 1.332\\ \hline

\end{tabular}
\vspace{2em}
\caption{Precision, Recall and loss values with edited loss-function ( without log-likelihood, only KL-divergence ) and original loss function ( KL-divergence with log-likelihood ). Original loss functions keeps track of the labels whereas with edited loss function, we have tried to capture the information retention of Knowledge Graphs in Language Models.}

\label{table:Results}

\end{center}
\end{table*}

Experiment results (see \textbf{Table} \ref{table:Results}) showed our hypothesis to be true. With the modified loss-function, the AUC-ROC and AUC-PR values dropped significantly. It was expected since we were not penalising wrong predictions anymore. Loss between attention layers also decreased as expected. The new loss was less than 10\% of the loss with original loss function. Still, the graph reproduced was not able to capture the underlying semantics of the input graph.



For generating Knowledge graphs we need an attention matrix. Once the attention matrix is generated, they are used with the existing pair of entities provided to find the path between those entities. The conditional probability of occurrence of each entity after a entity is used to determine the next entity. Once we have the path between two extreme entities, we combine them to define a sequence.

This entire process involves taking Knowledge Graph as an input feature $->$ Generating attention matrix using conditional probabilities $->$ Regeneration of Knowledge from trained attention matrix.

\section{Conclusion}
It is evidenced that due to the disparity in Knowledge Graphs (directed with relationships) and the fully connected graph interpretation of transformer architecture, graphical structure is imposed through KL divergence at all the layers, cannot capture the graphical structure  adequately. Thus,in cases when we are trying to predict relations from an attention matrix in the general case, it is quite possible that there are multiple relations. Thus taking only max attention value eliminates the other relations. An alternative could be to use a threshold value instead of max value, but choosing that threshold value is itself a challenging task. Also in case of incomplete data where there’s no KG triplet, it will tend to predict unwanted relations as the attention weights will still be significant. Thus the idea of graph structures from Language Models needs further research to overcome the limitations prevailing in the current models. Language models although excellent at GLUE tasks show significant limitations in semantic knowledge capture ultimately required for more intelligent systems, for example, to enable common sense reasoning.

\section{Future Work}

Knowledge-infused learning is an emerging field. A lot of research is going on to improve the existing models by infusing a knowledge base to provide better results.

Future research would be to restructure neurosymbolic approaches or graph neural network and similar approaches to construct a model which would be flexible with incorporating the richness of the graph structures. Constructing graphs from Language Models still requires a lot more research to work well with incomplete/extremely longer sentences and any other form of data which can’t be transformed to a sentence.

To summarize, here we have discussed the conclusions derived while formulating the idea. More research needs to be conducted in future to convert the idea into a working model. The direction would be to experiment with other models to further understand the boundaries of large transformer models and how they can be stretched.
 \newpage
\bibliographystyle{unsrtnat}
\bibliography{references}

\end{document}